\documentclass[fleqn,10pt]{wlscirep}
\usepackage[english]{babel}
\usepackage[utf8]{inputenc}
\usepackage[T1]{fontenc}
\usepackage{lineno}
\usepackage{graphicx} % Required for inserting images
\usepackage{longtable}

\title{Spatially Resolved Meteorological and Ancillary Data in Central Europe for Rainfall Streamflow Modeling}

\author[1,*]{Marc Aurel Vischer}
\author[1]{Noelia Otero}
\author[1,*]{Jackie Ma}
\affil[1]{Fraunhofer Heinrich-Hertz Institute, Applied Machine Learning Group, 10587 Berlin, Germany}

\affil[*]{corresponding authors: Marc Aurel Vischer (marc.aurel.vischer@hhi.fraunhofer.de), Jackie Ma (jackie.ma@hhi.fraunhofer.de)}

\begin{abstract}
We present a dataset for rainfall streamflow modeling that is fully spatially resolved with the aim of taking neural network-driven hydrological modeling beyond lumped catchments. To this end, we compiled data covering five river basins in central Europe: upper Danube, Elbe, Oder, Rhine, and Weser. The dataset contains meteorological forcings, as well as ancillary information on soil, rock, land cover, and orography. The data is harmonized to a regular $9km \times 9km$ grid and contains daily values that span from October 1981 to September 2011. We also provide code to further combine our dataset with publicly available river discharge data for end-to-end rainfall streamflow modeling.
\end{abstract}
\begin{document}

\flushbottom
\maketitle

\thispagestyle{empty}

\section*{Background \& Summary}

In recent years years, a substantial number of rainfall streamflow datasets were released that follow the example of the popular CAMELS dataset \cite{newmanDevelopment2015, addorCAMELS2017}. They cover Chile \cite{alvarez-garretonCAMELSCL2018}, Great Britain \cite{coxonCAMELSGB2020}, Brazil \cite{chagasCAMELSBR2020}, Australia \cite{fowlerCAMELSAUS2021}, the upper Danube basin \cite{klinglerLamaHCE2021}, France \cite{delaigueCAMELSFR2022}, Switzerland \cite{hogeCAMELSCH2023}, Denmark \cite{liuCAMELSDK2024} and Germany \cite{loritzCAMELSDE2024}. The publications of these datasets went hand in hand with a surge in popularity of neural network models for rainfall streamflow modeling, and their hunger for readily available, harmonized and tidy data. The central idea behind all these datasets is to leverage neural networks' flexibility to model hydrological processes not only from meteorological variables, but also consider additional static information such as land cover, soil and bedrock type and orographic features. Crucially, this data does not need to be cast into physical formulas, and neither do domain experts have to compile hydrological characteristics from them. Neural networks can extract relevant information from these data sources in a purely data-driven fashion, without ingesting additional domain expertise. Our choice of static information, also termed ancillary information, follows the groundbreaking work of Kratzert et al. \cite{kratzertImproved2019}. 
A common downside of all above-mentioned datasets is that they aggregate or \emph{lump} each variable within a catchment to a single value. By doing so, all information about spatial variability is lost: A pattern of soil types might be reduced to the most prevalent one, or a range of different temperatures might be averaged to a single mean value for a given day. This reduction of information is unnecessary and counter-intuitive, especially for large catchments with high spatial variability. The principle advantage of spatially resolved inputs is that they enable the model to capture spatial covariance among different variables, e.g. the interacting effects of soil sealing or steepness of terrain and a torrential rainfall.
Physical models, still the standard model type in active operation, resolve their equations on such a grid for exactly this reason. At the same time, training a neural network model benefits from vast amounts of data -  the more the better, as a general tendency. Additionally, as each point on the grid contains a complete, self-contained set of meteorological and ancillary variables, the grid locations can be processed independently. Neural networks are particularly efficient at such parallel processing of independent inputs. As a consequence, neural network models are capable of efficiently modeling hydrological processes in spatial detail even inside large basins. Recent advances in computer memory have made this kind of data processing practically feasible. 
With the publication of this dataset, we want to promote the development of neural network models beyond the scope of lumped catchments, closing one gap between them and state of the art operational physics-inspired models, and further improving their performance.
\\
We bundle 6 dynamic, meteorological features with 46 ancillary static features (3 hydrogeological features, 16 land cover features, 19 soil features and 8 orographical features). Our study area covers 5 basins in central Europe, namely the upper reaches of the Danube (until Bratislava), Elbe, Oder, Weser and Rhine. The dynamic data spans from 1st October 1981 to 30th September 2011. See figure \ref{fig:data_types} and tables \ref{tab:data_sources}, \ref{tab:dynamic_inputs}, \ref{tab:static_inputs} for details.
Along with the data, we release all scripts for processing the raw source data into the dataset that we provide. We also provide an additional script that combines our data with river discharge data after manual download from the the \href{https://portal.grdc.bafg.de/applications/public.html}{original provider}. This data can serve as targets for end-to-end training in data-driven rainfall streamflow modeling.

\section*{Methods}

\begin{figure}[h]
\centering
\includegraphics[width=\linewidth]{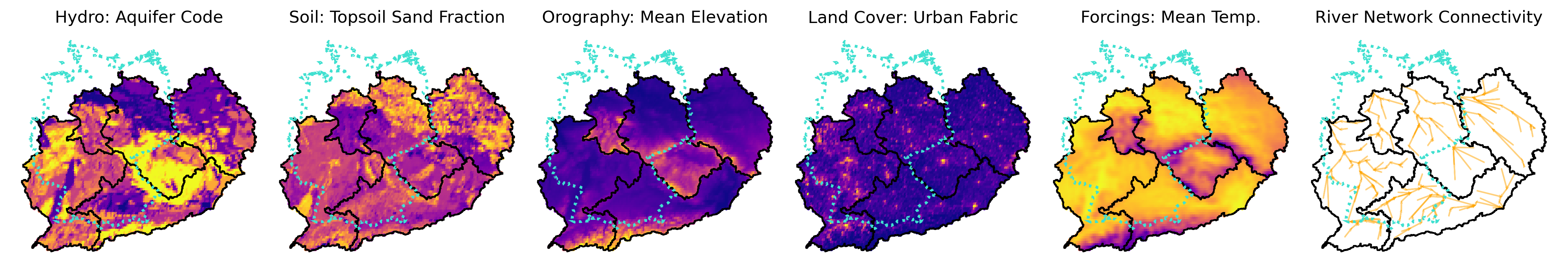}
\caption{Overview of study area and visualizations for an example feature of each type. The basins are outlined in black, with the outline of Germany \protect\footnotemark in turquoise for geographic reference. The right hand panel shows additional river network connectivity information as yellow arrows that can be derived from the GRDC data with code from our repository.}
\label{fig:data_types}
\end{figure}

Our dataset consists of data derived from a variety of publicly available sources - no new data was recorded. Our contribution consists in collecting the data and harmonizing it to a common grid for convenient model training. As smallest common denominator, we decided to use the grid of the ERA5-Land reanalysis dataset \cite{munozsabaterERA5Land2019, copernicusclimatechangeserviceERA5Land2022}. This dataset contains a vast number of meteorological variables, covering the entire world and resolved hourly from 1950 onwards. In this way, our dataset remains easily extendable, should a user like to e.g. include an additional meteorological variable in their experiments, extend the study area or increase the temporal resolution. Together with this spatiotemporal \emph{dynamical} data, we deliver various kinds of static or \emph{ancillary} features. The spatial data originally comes in different formats (vector or grid), projections and resolutions. All data sources were harmonized to the  grid of ERA5 by means of re-projecting and sub-sampling at the nodes of this grid. We thus created a dataset with a common two-dimensional, regular grid covering the earth's surface with a resolution of $0.1 ^{\circ} \times 0.1 ^{\circ}$ or roughly $9 km \times 9 km$. As a study area, we chose the entire river basins of Elbe, Oder, Weser and Rhine, as well as the upper reaches of the Danube basins, up to Bratislava. Together, these basins cover a contiguous 570.581 km² area of Germany and parts of neighboring countries, which we want to focus on in future work. Also, these basins are covered densely and uniformly with river gauging stations. Since this is not the case for the lower Danube basin however, we decided to only include part of the Danube basin. River discharge time series for the study area is available for download at the \href{https://portal.grdc.bafg.de/applications/public.html}{Global Runoff Data Center (GRDC) Data Portal} at daily resolution, to which we harmonized the inputs' temporal resolution. The dataset's temporal coverage is from 1st October 1980 to 30th September 2011, or in other words the 31 water years 1981 to 2011. Figure \ref{fig:data_types} gives an overview of the study area and visualizes an example feature for each of the five different sources of information contained in our dataset. The features are explained in the following subsections. Table \ref{tab:data_sources} contains references to the data sources, tables \ref{tab:dynamic_inputs} and \ref{tab:static_inputs} provide additional detail on our extracted features.

\footnotetext{The country boundary information was downloaded from \href{https://simplemaps.com/gis/country/de}{simplemaps}.}

\subsection*{Meteorological Forcings}
The meteorological forcings in our study were derived from the ERA5-Land dataset\footnote{The dataset was downloaded from the \href{https://cds.climate.copernicus.eu/datasets/reanalysis-era5-land?tab=overview}{Copernicus Climate Change Service (2022)}. The results contain modified Copernicus Climate Change Service information 2020. Neither the European Commission nor ECMWF is responsible for any use that may be made of the Copernicus information or data it contains.} \cite{munozsabaterERA5Land2019, copernicusclimatechangeserviceERA5Land2022}. Balancing costs and benefits, we downloaded the data every three hours, then aggregated each of the following variables daily: temperature two meters above surface was aggregated by calculating minimum, mean and maximum values; potential evapotranspiration was summed, and precipitation is aggregated by calculating sum and variance in order to capture how concentrated rainfall was over the course of the day. Table \ref{tab:dynamic_inputs} provides a summary of all dynamic variables.

\subsection*{Ancillary Data}

\emph{Hydrogeological properties} were derived from the International Hydrogeological Map of Europe (IHME) \footnote{\href{https://www.bgr.bund.de/EN/Themen/Wasser/Projekte/laufend/Beratung/Ihme1500/ihme1500_projektbeschr_en.html}{IHME1500 - Internationale Hydrogeologische Karte von Europa 1:1.500.000, version 1.2} © Bundesanstalt für Geowissenschaft und Rohstoffe, 2022.} \cite{guntherExtended}. The original dataset features six hydrogeological classes as well as two classes for snow-ice-fields and inland water bodies. The six classes represent the productivity of rock type, which indicates how easily water can dissipate through the bedrock. Classes are ordinal in that they are sorted by the corresponding productivity in ascending order. This allows us to take a non-rigorously defined but nonetheless informative average over the classes' proportions within each grid cell. We concatenate this productivity score with the binary categorical classes for snow-ice-fields and inland water bodies, each represented by a ratio of prevalence of this type of binary class within the grid cell.
\\
\emph{Land Cover information} was obtained from the Corine Land Cover Map\footnote{\href{https://land.copernicus.eu/en/products/corine-land-cover/clc-2012}{Corine Land Cover Map, version 2012}. Generated using European Union's Copernicus Land Monitoring Service information;  \url{https://doi.org/10.2909/916c0ee7-9711-4996-9876-95ea45ce1d27}. The Corine Land Cover Map data was created with funding by the European union. It was adapted and modified by the authors.} (CLC). This dataset classifies land cover at three different levels of detail, with increasingly differentiated (sub)classes. We decided to use the second level, which containing 16 classes in total. Similarly to the procedure applied to the hydrogeological properties, we calculated a distributional vector representing the proportion of a given class covering the grid cell.
\\
\emph{Soil type information} was obtained from the dataset European Soil Database Derived Data \footnote{\href{https://esdac.jrc.ec.europa.eu/content/european-soil-database-derived-data}{European Soil Database Derived Data}, created by the European Soil Data Centre with funding by the European union. It was adapted and modified by the authors. The authors' activities are not officially endorsed by the Union.} \cite{hiedererMapping2013, hiedererMapping2013a}. This dataset features 17 different physical properties, separately for top soil and lower soil. We calculate the average value of each feature within a grid cell.
\\
\emph{Orographic information} was derived from the European Union Digital Elevation Map\footnote{\href{https://sdi.eea.europa.eu/catalogue/srv/api/records/d08852bc-7b5f-4835-a776-08362e2fbf4b}{European Union Digital Elevation Map, version 1.1}. Generated using European Union's Copernicus Land Monitoring Service information. The European Union Digital Elevation Map created with funding by the European union. It was adapted and modified by the authors. The authors' activities are not officially endorsed by the Union.} (EU-DEM). Elevation was averaged within each grid cell, as well as the gradient in latitudinal and longitudinal direction, and the steepness or magnitude of the two-dimensional gradient. This yielded a total of four orographic features.
\\
Table \ref{tab:dynamic_inputs} provides an overview over all ancillary variables in the same ordering as we just introduced, which is also the ordering in the data file.

\section*{Data Records}

Dynamic meteorological forcing data and static ancillary data are stored in \href{https://doi.org/10.4211/hs.05d5633a413b4aec93b08a7e61a2abbb}{this hydroshare data repository} in separate NetCDF4 \cite{rewNetCDF42006} files. This format allows for named coordinates such as latitude and longitude or date for convenient selection on spatial and temporal domains, respectively. All variables are named in a self-explanatory manner and we provide labeled metadata. Tables \ref{tab:dynamic_inputs} and \ref{tab:static_inputs} provide a detailed overview of all features in the two files, Table \ref{tab:data_sources} lists their provenance.

\section*{Technical Validation}

All sources from which we obtained the original data have been widely used across various scientific fields for years, so we assume the original data to be valid. In order to technically validate our processing steps, we feature a testing script in our repository with extensive tests and visualizations of the compiled data. We also managed to successfully employ this dataset in training a neural network model for rainfall streamflow modeling (under review).

\section*{Usage Notes}

Along with the code to process the data, we provide a script that loads the data, selects subsets and visualizes them. This can serve as a starting point for the user to interact with the data. Furthermore, we provide code to wrap all the data in a PyTorch \cite{anselPyTorch2024} Dataset class for further processing.

\section*{Code availability}

The data was processed in several Python Jupyter Notebooks \cite{grangerJupyter2021} that can be found \href{https://gitlab.hhi.fraunhofer.de/vischer/spatial_streamflow_dataprep}{here}. The code requires Python 3.11 \cite{vanrossumPython2009} and is licensed under the Clear BSD licence. Additional dependencies are specified in an Anaconda \cite{Anaconda2020} environment specification contained in the repository. The scripts are stand-alone and do not require further input parameters.

\bibliography{DAKI-FWS}

\section*{Acknowledgements}

This work was supported by the Federal Ministry for Economic Affairs and Climate Action (BMWK) as grant DAKI-FWS (01MK21009A).

\section*{Author contributions statement}

M.A.V. compiled the data with crucial suggestions from N.F.O., processed the data, and wrote the manuscript with significant contributions from J.M.
All authors reviewed the manuscript. 

\section*{Competing interests}

The authors declare no competing interests.

\section*{Figures \& Tables}

\begin{table}[h]
  \resizebox{\textwidth}{!}{
    \begin{tabular}{llll}
        Type & Dataset & Author & Citation \\ \hline
        \emph{Forcings / Dynamic Inputs} & ~ & ~ & ~ \\ \hline
        Meteorological Variables & \href{https://cds.climate.copernicus.eu/datasets/reanalysis-era5-land?tab=overview}{ERA5-Land} & \href{https://climate.copernicus.eu/}{Copernicus Climate Change Service (CCCS)} & \cite{munozsabaterERA5Land2019, copernicusclimatechangeserviceERA5Land2022} \\ \hline
        \emph{Ancillary Data / Static Inputs} & ~ & ~ & ~ \\ \hline
        Hydrogeological Properties & \href{https://www.bgr.bund.de/EN/Themen/Wasser/Projekte/laufend/Beratung/Ihme1500/ihme1500_projektbeschr_en.html}{IHME hydrogeological map v1.2 in vector data format} & \href{https://www.bgr.bund.de/EN/Home/homepage_node_en.html}{German Federal Institute for Geosciences and Natural Resources (BGR)} & \cite{guntherExtended} \\ \hline
        Land Cover & \href{https://land.copernicus.eu/en/products/corine-land-cover/clc-2012}{Corine Land Cover Map, version 2012} & \href{https://land.copernicus.eu/en}{Copernicus Land Monitoring Service (CLMS)} & ~ \\ \hline
        Soil Type (Top and Lower Soil) & \href{https://esdac.jrc.ec.europa.eu/content/european-soil-database-derived-data}{European Soil Database Derived Data} & \href{https://esdac.jrc.ec.europa.eu/}{European Soil Data Centre (ESDAC)} & \cite{hiedererMapping2013, hiedererMapping2013a} \\ \hline
        Orographic & \href{https://sdi.eea.europa.eu/catalogue/srv/api/records/3473589f-0854-4601-919e-2e7dd172ff50}{European Union Digital Elevation Map (EU-DEM), version 1.1} & \href{https://land.copernicus.eu/en}{Copernicus Land Monitoring Service (CLMS)} & ~ \\ \hline
    \end{tabular}
  }
\caption{\label{tab:data_sources} Overview of source datasets and their authors for dynamic data / meteorological forcings contained in file \emph{forcings\_publish.nc} and static / ancillary data contained in \emph{ancillary\_publish.nc}. See tables \ref{tab:dynamic_inputs} and \ref{tab:static_inputs} for more details on derived features.}
\end{table}

\begin{table}[h]
  \resizebox{\textwidth}{!}{
    \begin{tabular}{p{.03\textwidth} p{.33\textwidth} p{.28\textwidth}  p{.08\textwidth}  p{.16\textwidth} }
        Index & Name & Feature & Origin & Aggregation \\ \hline
        00 & t2m\_min & Temperature 2m above ground & ERA5 & Daily Minimum \\ \hline
        01 & t2m\_mean & ~ & ~ & Daily Mean \\ \hline
        02 & t2m\_max & ~ & ~ & Daily Maximum \\ \hline
        03 & pev & Potential evapotranspiration & ~ & Daily Sum \\ \hline
        04 & tp\_sum & Precipitation & ~ & Daily Sum \\ \hline
        05 & tp\_var & ~ & ~ & Daily Variance \\  \hline
    \end{tabular}
  }  
\caption{\label{tab:dynamic_inputs} Overview of dynamic input features in the file \emph{forcings\_publish.nc}. Empty cells indicate that the value is identical to the one above. Each of these features is a two dimensional array with grid cell ID and date as indices. The file also provides longitude and latitude coordinates on the grid cell index dimension for convenient selection.}
\end{table}

\begin{table}[h]
  \resizebox{\textwidth}{!}{
        \begin{tabular}{p{.03\textwidth} p{.33\textwidth} p{.28\textwidth}  p{.08\textwidth}  p{.16\textwidth} }
            Index & Name & Feature & Origin & Aggregation \\ \hline
            00 & IHME\_AQUIF\_CODE & Rock Productivity & IHME & Averaged Classes \\ \hline
            01 & IHME\_INLAND\_WATER & Inland Water Body & ~ & Fraction \\ \hline
            02 & IHME\_SNOW\_ICE\_FIELD & Permanent Snow-Ice Field & ~ & ~ \\ \hline
            03 & CLC\_11\_Artificial\_surfaces\_Urban\_fabric & ~ & CLC & ~ \\ \hline
            04 & CLC\_12\_Artificial\_surfaces\_Industrial,\_commercial\_and\_transport\_units & ~ & ~ & ~ \\ \hline
            05 & CLC\_13\_Artificial\_surfaces\_Mine,\_dump\_and\_construction\_sites & ~ & ~ & ~ \\ \hline
            06 & CLC\_14\_Artificial\_surfaces\_Artificial,\_non\_agricultural\_vegetated\_areas & ~ & ~ & ~ \\ \hline
            07 & CLC\_21\_Agricultural\_areas\_Arable\_land & ~ & ~ & ~ \\ \hline
            08 & CLC\_22\_Agricultural\_areas\_Permanent\_crops & ~ & ~ & ~ \\ \hline
            09 & CLC\_23\_Agricultural\_areas\_Pastures & ~ & ~ & ~ \\ \hline
            10 & CLC\_24\_Agricultural\_areas\_Heterogeneous\_agricultural\_areas & ~ & ~ & ~ \\ \hline
            11 & CLC\_31\_Forest\_and\_seminatural\_areas\_Forest & ~ & ~ & ~ \\ \hline
            12 & CLC\_32\_Forest\_and\_seminatural\_areas\_Shrub\_and\_or\_herbaceous\_vegetation\_associations & ~ & ~ & ~ \\ \hline
            13 & CLC\_33\_Forest\_and\_seminatural\_areas\_Open\_spaces\_with\_little\_or\_no\_vegetation\_ & ~ & ~ & ~ \\ \hline
            14 & CLC\_41\_Wetlands\_Inland\_wetlands & ~ & ~ & ~ \\ \hline
            15 & CLC\_42\_Wetlands\_Coastal\_wetlands & ~ & ~ & ~ \\ \hline
            16 & CLC\_51\_Water\_bodies\_Inland\_waters & ~ & ~ & ~ \\ \hline
            17 & CLC\_51\_Water\_bodies\_Marine\_waters & ~ & ~ & ~ \\ \hline
            18 & CLC\_No\_data & ~ & ~ & ~ \\ \hline
            19 & SOIL\_STU\_EU\_S\_SILT & Subsoil: Silt Content & ESDAC & Arithmetic Mean \\ \hline
            20 & SOIL\_STU\_EU\_T\_SAND & Topsoil: Sand Content & ~ & ~ \\ \hline
            21 & SOIL\_SMU\_EU\_S\_TAWC & \mbox{Subsoil: Total Available Water Content from Pedotransfer Rule} & ~ & ~ \\ \hline
            22 & SOIL\_SMU\_EU\_T\_TAWC & \mbox{Topsoil: Total Available Water Content from Pedotransfer Rule} & ~ & ~ \\ \hline
            23 & SOIL\_STU\_EU\_T\_BD & Topsoil: Bulk Density & ~ & ~ \\ \hline
            24 & SOIL\_STU\_EU\_T\_TAWC & \mbox{Topsoil: Total Available Water Content from Pedotransfer Function} & ~ & ~ \\ \hline
            25 & SOIL\_STU\_EU\_S\_GRAVEL & Subsoil: Coarse Fragments & ~ & ~ \\ \hline
            26 & SOIL\_STU\_EU\_DEPTH\_ROOTS & Depth Available to Roots & ~ & ~ \\ \hline
            27 & SOIL\_STU\_EU\_T\_GRAVEL & Topsoil: Coarse Fragments & ~ & ~ \\ \hline
            28 & SOIL\_STU\_EU\_S\_TEXT\_CLS & Subsoil: Texture Class & ~ & ~ \\ \hline
            29 & SOIL\_STU\_EU\_T\_OC & Topsoil: Organic Content & ~ & ~ \\ \hline
            30 & SOIL\_STU\_EU\_S\_SAND & Subsoil: Sand Content & ~ & ~ \\ \hline
            31 & SOIL\_STU\_EU\_T\_CLAY & Topsoil: Clay Content & ~ & ~ \\ \hline
            32 & SOIL\_STU\_EU\_T\_TEXT\_CLS & Topsoil: Texture Class & ~ & ~ \\ \hline
            33 & SOIL\_STU\_EU\_T\_SILT & Topsoil: Silt Content & ~ & ~ \\ \hline
            34 & SOIL\_STU\_EU\_S\_BD & Subsoil: Bulk Density & ~ & ~ \\ \hline
            35 & SOIL\_STU\_EU\_S\_TAWC & \mbox{Subsoil: Total Available Water Content from Pedotransfer Function} & ~ & ~ \\ \hline
            36 & SOIL\_STU\_EU\_S\_OC & Subsoil: Organic Carbon Content & ~ & ~ \\ \hline
            37 & SOIL\_STU\_EU\_S\_CLAY & Subsoil: Clay Content & ~ & ~ \\ \hline
            38 & DEM\_elevation\_mean & ~ & EU-DEM & ~ \\ \hline
            39 & DEM\_grad\_x\_mean & ~ & ~ & ~ \\ \hline
            40 & DEM\_grad\_y\_mean & ~ & ~ & ~ \\ \hline
            41 & DEM\_steepness\_mean & ~ & ~ & ~ \\ \hline
            42 & DEM\_elevation\_std & ~ & ~ & Standard Deviation \\ \hline
            43 & DEM\_grad\_x\_std & ~ & ~ & ~ \\ \hline
            44 & DEM\_grad\_y\_std & ~ & ~ & ~ \\ \hline
            45 & DEM\_steepness\_std & ~ & ~ & ~ \\ \hline
        \end{tabular}
    }
\caption{\label{tab:static_inputs} Overview of static input features in the file \emph{ancillary\_publish.nc}. Empty cells indicate that the value is identical to the one above. Explanations of the features derived from CLC and elevation map were omitted because the names are self-explanatory. Each of these features is a one dimensional array with grid cell ID as index. The file also provides longitude and latitude coordinates on the index dimension for convenient selection.}
\end{table}

\end{document}